\documentclass{ecai2014}
\usepackage[utf8]{inputenc}

\usepackage{microtype}

\newif\ifdraft\drafttrue
\newif\iffinal\finalfalse
\newif\ifdotikz\dotikzfalse

\draftfalse
\finaltrue

\usepackage{times}
\usepackage{graphicx}
\graphicspath{{figures/}}
\usepackage{latexsym}
\usepackage{amsmath}
\usepackage{amstext}
\usepackage{amssymb}
\usepackage{amsfonts}
\usepackage{verbatim}
\usepackage{color}
\usepackage{paralist}
\usepackage{enumerate}
\usepackage{ifpdf}
\usepackage[amsmath,thmmarks]{ntheorem}
\newtheorem{definition}{Definition}
{\theorembodyfont{\normalfont}
 \theoremsymbol{\ensuremath{\medblacksquare}}
 \newtheorem{example}{Example}
}
{\theorembodyfont{\normalfont}
 \theoremheaderfont{\normalfont\itshape}
 \theoremseparator{.~}
 \theoremsymbol{\ensuremath{\Box}}
 
}

\newcommand{\entails}{\Vdash}
\newcommand{\notentails}{\nVdash}

\newcommand{\smallmathnl}{\ensuremath{\\[-0.7ex]}}

\newcommand{\leanparagraph}[1]{\smallskip\noindent\textbf{#1.} }

\newcommand*{\medblacksquare}{\mathbin{\scalebox{0.6}{\ensuremath{\blacksquare}}}} %

\newcommand{\nop}[1]{}
\newcommand{\todo}[1]{\begin{scriptsize}\textbf{TODO #1}\end{scriptsize}}

\newcommand{\mb}[1]{\ensuremath{\mathbf{#1}}}

\newcommand{\Predicates}{\ensuremath{\cP}}
\newcommand{\Constants}{\ensuremath{\cC}}
\newcommand{\Variables}{\ensuremath{\cV}}
\newcommand{\Terms}{\ensuremath{\cT}}
\newcommand{\Atoms}{\ensuremath{\cA}}
\newcommand{\GroundAtoms}{\ensuremath{\cG}}
\newcommand{\Formulas}{\ensuremath{\cF}}
\newcommand{\TimeVariables}{\ensuremath{\cU}}

\newcommand{\tup}[1]{\ensuremath{\langle #1 \rangle}}

\newcommand{\restr}[2]{\ensuremath{#1}|_{#2}}

\newcommand{\card}[1]{\ensuremath{\# #1}}

\newcommand{\newquery}[2]{\ensuremath{#1\lbrack #2 \rbrack}}

\newcommand{\tram}{\ensuremath{\mathit{tr}}}
\newcommand{\bus}{\ensuremath{\mathit{bus}}}

\mathchardef\minus="2D

\makeatletter
\newcommand{\dotminus}{\mathbin{\text{\@dotminus}}}
\newcommand{\@dotminus}{%
  \ooalign{\hidewidth\raise1ex\hbox{.}\hidewidth\cr$\m@th-$\cr}%
}
\makeatother
\newcommand{\intpr}{\upsilon} %

\DeclareMathOperator{\varAs}{\sigma}
\DeclareMathOperator{\timeAs}{\tau}

\DeclareMathOperator{\querySubst}{\Theta}

\DeclareMathOperator{\idx}{idx}

\newcommand{\window}{\ensuremath{\boxplus}}
\newcommand{\windowf}{\ensuremath{\hat{W}}}
\newcommand{\extwf}{\ensuremath{\hat{w}}}

\newcommand{\yes}{\ensuremath{\mathit{yes}}}
\newcommand{\no}{\ensuremath{\mathit{no}}}

\newcommand{\cA}{\ensuremath{\mathcal{A}}}

\newcommand{\cC}{\ensuremath{\mathcal{C}}}

\newcommand{\cF}{\ensuremath{\mathcal{F}}}
\newcommand{\cG}{\ensuremath{\mathcal{G}}}

\newcommand{\cP}{\ensuremath{\mathcal{P}}}

\newcommand{\cS}{\ensuremath{\mathcal{S}}}
\newcommand{\cT}{\ensuremath{\mathcal{T}}}
\newcommand{\cU}{\ensuremath{\mathcal{U}}}
\newcommand{\cV}{\ensuremath{\mathcal{V}}}

\newcommand{\bbN}{\ensuremath{\mathbb{N}}}

\newcommand{\ssS}{\ensuremath{{\scriptscriptstyle S}}}

\ifdotikz
\usepackage{pgf}
\usepackage{tikz}
\usetikzlibrary{arrows,positioning,automata,decorations,fit,backgrounds}
\usepgflibrary{shapes.geometric} %
\usetikzlibrary{shapes.geometric} %
\usepgflibrary{decorations.pathmorphing} %
\usetikzlibrary{decorations.pathmorphing} %
\usepgflibrary{decorations.text} %
\usetikzlibrary{decorations.text} %
\pgfrealjobname{reactknow2014}
\else
\ifpdf
\long\def\beginpgfgraphicnamed#1#2\endpgfgraphicnamed{\includegraphics{#1}}
\else
\usepackage{epsfig}
\long\def\beginpgfgraphicnamed#1#2\endpgfgraphicnamed{\epsfig{file=#1.eps}}
\fi
\fi

\begin{document}

\title{\hspace{-0.059cm}Towards Ideal Semantics for Analyzing Stream
  Reasoning%
\thanks{Supported by the Austrian Science Fund (FWF) project 26471.}
}

\author{Harald Beck \and Minh Dao-Tran \and Thomas Eiter \and Michael Fink
  \institute{Institut f\"ur Informationssysteme, Technische Universit\"at Wien.
    email: \{beck,dao,eiter,fink\}@kr.tuwien.ac.at} }

\maketitle
\bibliographystyle{ecai2014}

\begin{abstract}%
  The rise of smart applications has drawn interest to logical reasoning
  over data streams. Recently, different query languages and stream
  processing/reasoning engines were proposed in different communities. However, due
  to a lack of theoretical foundations, the expressivity and semantics of these diverse
  approaches are given only informally. Towards clear specifications and
  means for analytic study, a formal framework is needed to define their
  semantics in precise terms.
  To this end, we present a first step towards an ideal semantics that
  allows for exact descriptions and comparisons of stream
  reasoning systems.
\end{abstract}
\section{Introduction \ifdraft(1 - 1.5p)\fi}
\label{sec:intro}
The emergence of sensors, networks, and mobile devices has generated a
trend towards \emph{pushing} rather than \emph{pulling} of data in
information processing. In the setting of \emph{stream
  processing}~\cite{BabuW01} studied by the database community, input
tuples dynamically arrive at the processing systems in form of possibly
infinite streams. To deal with unboundedness of data, such systems
typically apply %
\emph{window operators} to obtain snapshots of recent data.
The user then runs \emph{continuous queries} which are either periodically driven by time
or eagerly driven by the arrival of new input.
The Continuous Query Language~(CQL)~\cite{ArasuBW06} is a well-known
stream processing language. It has a syntax close to SQL and a clear
operational semantics.

\begin{figure}[t]
  \centering
  \footnotesize
  \beginpgfgraphicnamed{traffic-scenario}
  \begin{tikzpicture}[scale=0.55,node distance=0.4cm,>=stealth']
    \node (tram_a_p1_2) at (2,1) {$\tram(a,p_1)$};
    \node (bus_c_p1_2) [above of=tram_a_p1_2] {$\bus(c,p_1)$};
    \node (tram_d_p2_8) at (8,1) {$\tram(d,p_2)$};
    \node (bus_e_p2_11) at (11,1) {$\bus(e,p_2)$};
    \draw [->] (0,0) -- (13,0);

    \foreach \i in {0,2,8,11}
    {
      \draw (\i cm,5pt) -- (\i cm,-5pt) node[anchor=north] {$\i$};
    }
  \end{tikzpicture}
  \endpgfgraphicnamed
  \caption{Traffic scenario with arrivals of trams and buses}
  \label{fig:traffic-scenario}
\end{figure}
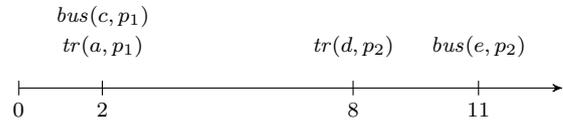

Recently, the rise of \emph{smart applications} such as smart
cities, smart home, smart grid, etc., has raised interest in
the topic of \emph{stream reasoning}~\cite{VCHF09}, i.e.,
logical reasoning on streaming data.
To illustrate our contributions on this topic, we use an example from the
public transport domain.
\begin{example}
  \label{ex:traffic-scenario}
  To monitor a city's public transportation, the city traffic center
  receives sensor data at every stop regarding tram/bus appearances
  of the form~${\tram(X,P)}$ and~${\bus(X,P)}$ where~$X$,~$P$ hold the
  tram/bus and stop identifiers, respectively. On top of this
  streaming data tuples (or atoms), one may ask different queries, e.g.,
  to monitor the status of the public transport system. %
  To keep things simple, we start with stream processing queries:
  \begin{enumerate}[($q_1$)]
  \item\label{q1} At stop~$P$, did a tram and a bus arrive within the
    last~$5$ min?
  \item\label{q2} At stop~$P$, did a tram and a bus arrive \emph{at the same time} within the last~$5$ min?
  \end{enumerate}
  Consider the scenario of Fig.~\ref{fig:traffic-scenario} which depicts
  arrival times of trams and buses.
  The answer to query~($q_2$) is yes for stop~$p_2$ and all
  time points from~$2$ to~$7$. Query ($q_1$) also succeeds for~$p_1$
  from~$11$ to~$13$.

  As for stream reasoning, later we will additionally consider a more involved query,
  where we are
  interested in whether a bus always arrived within three minutes after
  the last two arrivals of trams.
\end{example}
Different communities have contributed to different aspects of this
topic.
\begin{inparaenum}[(i)]
\item\label{sw-community} The Semantic Web community extends SPARQL to
  allow querying on streams of RDF triples. %
  Engines such as CQELS~\cite{PhuocDPH11} and~C-SPARQL~\cite{BarbieriBCVG10} also follow the snapshot semantics
  approach of~CQL.
\item\label{krr-community} In Knowledge Representation and Reasoning
  (KRR), first attempts towards expressive stream reasoning
  have been carried out by considering continuous data in Answer Set
  Programming (ASP)~\cite{DoLL11,GebserGKOSS2012} or extending Datalog
  to sequential logic programs~\cite{Zaniolo12}.
\end{inparaenum}
However, the state of the art in either field has several shortcomings.

Approaches in~\eqref{sw-community} face difficulties with extensions of
the formalism to incorporate the Closed World Assumption,
nonmonotonicity, or non-determinism. Such features are important to deal
with missing of incomplete data, which can temporarily happen
due to unstable network connections or hardware failure. In this case,
engines like \mbox{C-SPARQL} and CQELS remain idle, while some output based on
default reasoning might be useful.
Moreover, e.g., in the use case of dynamic planning on live data, multiple
plans shall be generated based on previous choices and the availability
of new data.
This is not possible with current
deterministic approaches.

On the other hand, advanced reasoning has extensively been investigated
in~\eqref{krr-community} but traditionally only on static data. First
attempts towards stream reasoning reveal many problems to solve. The
plain approach of~\cite{DoLL11} periodically calls the dlvhex
solver~\cite{EiterIST06} but is not capable of incremental reasoning and
thus fails under heavy load of data.
\mbox{StreamLog}~\cite{Zaniolo12} is an extension of Datalog towards
stream reasoning. It always computes a single model and does not
consider windows. Time-decaying logic
programs~\cite{GebserGKOSS2012} attempt to implement time-based
windows in reactive ASP~\cite{GebserKKOST08} but the relation to other
stream processing/reasoning approaches has not yet been explored.

Moreover, as observed in~\cite{DindarTMHB13}, conceptually identical queries may
produce different results in different engines. While such deviations
may occur due to differences~(i.e., flaws) in implementations of a
common %
semantics, they might also arise from (correct
implementations of) different semantics.
For a user it is important to know the exact capabilities and the
semantic behavior of a given approach.
However, %
there is a lack of theoretical underpinning %
or a formal framework
for stream
reasoning %
that allows to capture different (intended) semantics in precise terms.
Investigations of specific languages, as well as comparisons between
different approaches, %
are confined to experimental analysis~\cite{ldpbef12}, or informal examination
on specific examples.
A systematic investigation, however, requires a formalism %
to rigorously describe the expressivity and the properties of a language.

\leanparagraph{Contributions}
We present a first step towards a \emph{formal framework for stream reasoning} that~%
\begin{inparaenum}[(i)]
\item\label{contribs-1} provides a common ground to express concepts
  from different stream processing/reasoning formalisms and engines;~%
\item\label{contribs-2} allows systematic analysis and comparison
  between existing stream processing/reasoning semantics; and~%
\item\label{contribs-3} also provides a basis for extension towards more
  expressive stream reasoning. Moreover, we present~%
\item\label{contribs-4} exemplary formalizations based on a running
  example, and~%
\item\label{contribs-5} compare our approach to existing work.
\end{inparaenum}

Thereby, we aim at capturing idealized %
stream reasoning semantics
where %
no information is dropped
and semantics are characterized as providing an abstract view over the entire
stream. %
Second, we idealize with respect
to implementations and do not consider processing time, delays or
outages in the semantics itself.
Moreover, we allow for a high degree of expressivity %
regarding
time reference: %
We distinguish notions of truth of a formula~%
\begin{inparaenum}[(i)]
\item at specific time points,~%
\item some time point within a window, or~%
\item all time points in a window. Moreover, we allow~%
\item for nested window operators,
\end{inparaenum}
which provide a means to reason over streams within the language itself (a formal counterpart to repeated runs of continuous queries).
\section{Streams \ifdraft(1 - 1.5p)\fi}
\label{sec:streams}
In this section, we introduce a logic-oriented view of streams and
formally define generalized versions of prominent window functions.
\subsection{Streaming Data}

A stream is usually seen as a sequence, set or bag of tuples with a
timestamp.
Here, we view streams as functions from a discrete time domain
to sets of logical atoms and
assume no fixed schema for tuples.

We build upon mutually disjoint sets of predicates~$\Predicates$,
constants~$\Constants$, variables~$\Variables$ and time
variables~$\TimeVariables$. The set $\Terms$ of terms is given
by ${\Constants \cup \Variables}$ and the set~$\Atoms$ of atoms is
defined as ${\{p(t_1,\dots,t_n) \mid p \in \Predicates,\, t_1,\dots,t_n
  \in \Terms\}}$. The set $\GroundAtoms$ of \emph{ground atoms} contains
all atoms~${p(t_1,\dots,t_n) \in \cA}$ such that ${\{t_1,\dots,t_n\}
  \subseteq \Constants}$.
If~${i,j \in \bbN}$, the set~${[i,j] = \{ k\! \in\! \bbN \mid i \leq k \leq
  j\}}$ is called an \emph{interval}. %
\begin{definition}[Stream]\label{def:stream}
  Let~$T$ be an interval and~${\intpr\colon \bbN \rightarrow
    2^\GroundAtoms}$ an \emph{interpretation function} such
  that~${\intpr(t) = \emptyset}$ for all~${t \in \bbN \setminus
    T}$. Then, the pair~${S=(T,\intpr)}$ is called a \emph{stream},
  and~$T$ is called the \emph{timeline} of~$S$.
\end{definition}
The elements of a timeline are called~\emph{time points}
or~\emph{time\-stamps}.  A stream~${S'=(T',\intpr')}$ is a
\emph{substream} or \emph{window} of stream~${S=(T,\intpr)}$,
denoted~${S' \subseteq S}$, if~${T' \subseteq T}$ and~${\intpr'(t')
  \subseteq \intpr(t')}$ for all~${t' \in T'}$.
The~\emph{cardinality} of~$S$, denoted~$\card{S}$, is defined
by~$\Sigma_{t\in T}|\intpr(t)|$.
The \emph{restriction of~$S$ to~${T' \subseteq T}$},
denoted ${\restr{S}{T'}}$,
is the
stream~${(T',\restr{\intpr}{T'})}$, where~${\restr{\intpr}{T'}}$ is the
usual domain restriction of function~$\intpr$.
\begin{example}[cont'd]
  \label{ex:stream}
  The input for the scenario in Example~\ref{ex:traffic-scenario} can be
  modeled as a stream~$S=(T,\intpr)$ where~$T=[0,13]$ and
  \[
  \begin{array}{l@{~=~}l@{\qquad}l@{~=~}l}
    \intpr(2)  & \{\tram(a,p_1),\bus(c,p_1)\} & \intpr(11)  & \{\bus(e,p_2)\} \\
    \intpr(8)  & \{\tram(d,p_2)\}             & \intpr(t) & \emptyset~~\text{otherwise.}
  \end{array}
  \]
  The interpretation~$\intpr$ can be equally represented as the
  following set:\\
  \mbox{$\{2 \!\mapsto\! \{\tram(a,p_1),\!\bus(c,p_1)\},
    8\!\mapsto\!\{\tram(d,p_2)\}, 11\!\mapsto\!\{\bus(e,p_2)\}\!\}~$}
\end{example}
\subsection{Windows}
\label{sec:windows}
An essential aspect of stream reasoning is to limit the considered data
to so-called \emph{windows}, i.e., recent substreams, in order to limit
the amount of data and forget outdated information.
\begin{definition}[Window function]\label{def:window-function}
  A \emph{window function} maps from a stream~${S=(T,\intpr)}$ and a
  time point~${t \in T}$ to a window~${S' \subseteq S}$.
\end{definition}
The usual time-based window of size~$\ell$ \cite{ArasuBW06} contains
only the tuples of the last~$\ell$ time units. We give a generalized
definition where the window can also include the tuples of the
future~$u$ time points. Based on query time~$t$ and a step size~$d$, we
derive a \emph{pivot} point~$t'$ from which an interval~${[t_\ell,t_u]}$
is selected by looking backward (resp., forward)~$\ell$
(resp.,~$u$) time units from~$t'$, i.e.,~${t_\ell + \ell = t'}$ and~${t'
  + u = t_u}$.
\begin{definition}[Time-based window]\label{def:time-based-window}
Let~${S=(T,\intpr)}$ be a stream with timeline~${T=[t_{min},t_{max}]}$,
let ${t \in T}$,  and let ${d,\ell,u \in \bbN}$ such that~${d \leq \ell
+ u}$.
  The \emph{time-based window with range~$(\ell,u)$ and step size~${d}$
    of~$S$ at time~${t}$} is defined by
\[w^{\ell,u}_d(S,t)=(T',\restr{\intpr}{T'}),\]
where $T'= \lbrack t_\ell, t_u \rbrack$,
${t_\ell=\max \{t_{min},t'-\ell\}}$ with $t' = \lfloor \frac{t}{d}
\rfloor \cdot d$, and~$t_u=\min\{t'+u,t_{max}\}$.
\end{definition}
For time-based windows that target only the past~$\ell$ time points, we
abbreviate~$w^{\ell,0}_d$ with~$w^\ell_d$. For windows which target only
the future, we write~$w^{+u}_d$ for~$w^{0,u}_d$. If the step size~$d$ is
omitted, we take~${d=1}$. Thus, the standard sliding window with
range~$\ell$ is
denoted by~$w^\ell$. %

The CQL~\cite{ArasuBW06} syntax for~$w^{\ell}_d$ is~\texttt{[Range l
  Slide d]} and~$w^\ell$ corresponds to~\texttt{[Range l]}. Moreover,
the window~\texttt{[Now]} equals~\texttt{[Range 0]} and thus corresponds
to~$w^0$. The entire past stream, selected by~\texttt{[Range
  Unbounded]} in CQL, is obtained by~$w^t$, where~$t$ is the query
time.
To consider the entire stream (including the future), we can use~$w^n$,
where~${n=\max T}$.

Furthermore, we obtain \emph{tumbling windows} by setting~${d=\ell +
  u}$.

\begin{figure}[t]
  \centering
  \beginpgfgraphicnamed{generalized-time-based-windows}
  \begin{tikzpicture}[scale=0.8,node distance=0.4cm,>=stealth']
    \draw [->] (0,0) -- (10,0);

    \foreach \i in {0,...,9}
    {
      \draw (\i cm,-5pt) -- (\i cm,5pt) node[anchor=south] {\scriptsize $\i$};
      \draw [densely dotted] (\i,0) -- (\i,-4.1);
    }

    \foreach \xleft/\xright/\ytop/\ybot/\tprime in {0/1/-0.3/-1.1/0,1/4/-1.4/-2.2/3,4/7/-2.5/-3.3/6}{
      \draw [densely dashed] (\xleft,\ytop) --
                             (\xright,\ytop) --
                             (\xright,\ybot) --
                             (\xleft,\ybot) --
                             cycle;

      \foreach \ystep in {0.2,0.4,0.6}{
        \node at (\tprime,\ytop-\ystep) {\tiny $\times$};
      }

      \foreach \xstep/\ystep in {0/0.2,1/0.4,2/0.6}{
        \node at (\tprime+\xstep,\ytop-\ystep) {\tiny $\bullet$};
      }
    }

    \draw [<->] (4,-3.9) -- node[anchor=south] {\scriptsize $\ell$} (6,-3.9);
    \draw [<->] (6,-3.9) -- node[anchor=south] {\scriptsize $u$} (7,-3.9);

    \node at (2.7,-4.6) {\scriptsize $\bullet\colon$ query times $t$};
    \node at (6.3,-4.6) {\scriptsize $\times\colon$ pivot points $t'$};
  \end{tikzpicture}
  \endpgfgraphicnamed
  \caption{Time-based window $w_3^{2,1}$ with range~$(2,1)$ and step
    size~$3$}
  \label{fig:generalized-time-based-windows}
\end{figure}
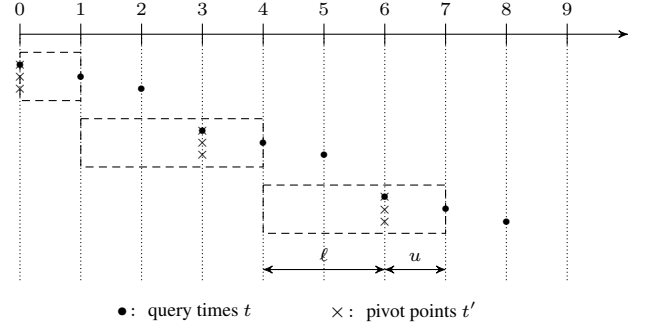

\begin{example}[cont'd]
  \label{ex:time-based-windows}
  To formulate the monitoring over the stream~$S$ of
  Example~\ref{ex:stream}, one can use a time-based window~$w^5$ with a
  range of~$5$ minutes (to the past) and step size of~$1$ minute, i.e.,
  the granularity of~$T$. The results of applying this window function
  at~${t=5,11}$ are
  \begin{align*}
    w^5(S,5) & = ([0,5],\{2\mapsto\{\tram(a,p_1),\bus(c,p_1)\}\})\text{,
      and}\smallmathnl
    w^5(S,11)& =
    ([6,11],\{8\mapsto\{\tram(d,p_2)\},11\mapsto\{\bus(e,p_2)\}\}).
  \end{align*}
  Moreover, consider a time-based (tumbling) window with range~$(2,1)$
  and step size~$3$.
  For~${t_1=5}$, we have~${t_1' = \lfloor \frac{5}{3} \rfloor \cdot
    3=3}$,
  thus~${T'_1=[\max\{0,3-2\},\min\{3+1,13\}]=[1,4]}$. For~${t_2=11}$, we
  get~${t_2'=9}$ and~${T'_2=[7,10]}$.
  The windows for~${t=5,11}$ are
  \begin{align*}
    w^{2,1}_3(S,5) & = ([1,4],\{2\mapsto\{\tram(a,p_1),\bus(c,p_1)\}\})\text{,
      and}\smallmathnl
    w^{2,1}_3(S,11)& = ([7,10],\{8\mapsto\{\tram(d,p_2))\}\}).
  \end{align*}
  Figure~\ref{fig:generalized-time-based-windows} illustrates the
  progression of this window with time.
\end{example}

The goal of the standard tuple-based window with count~$n$ is to fetch
the most recent~$n$ tuples. Again, we give a more general definition
which may consider future tuples. That is, relative to a time point~${t
  \in T=[t_{min},t_{max}]}$, we want to obtain the most recent~$\ell$
tuples (of the past) and next~$u$ tuples in the future. Thus, we must
return the stream restricted to the smallest interval~${T'=[t_\ell,t_u]
  \subseteq T}$, where~${t_\ell \leq t \leq t_u}$, such that~$S$
contains~$\ell$ tuples in the interval~${[t_\ell,t]}$ and~$u$ tuples in
the interval~${[t+1,t_u]}$. In general, we have to discard tuples
arbitrarily at time points~$t_\ell$ and~$t_u$ in order to receive
\emph{exactly}~$\ell$ and~$u$ tuples, respectively. In extreme cases,
where fewer than~$\ell$ tuples exist in~$[t_{min},t]$, respectively
fewer than~$u$ tuples in~$[t+1,t_{max}]$, we return all tuples of the
according intervals. Given~${t \in T}$ and the tuple counts~${\ell, u
  \in \bbN}$, we define the \emph{tuple time bounds}~$t_\ell$ and~$t_u$
as

{\small
\[
\begin{array}{l@{\,}l@{}l}
  t_\ell &= \max\, &\{t_{min}\} \cup \{ {t' \!\mid t_{min} \leq t' \leq
    t} \,\land\, {\card{(\restr{S}{[t',t]})}\! \geq\! \ell} \},~\text{and}\\
  t_u &= \,\min\, &\{t_{\max}\} \cup \{ {t' \!\mid t\! +\! 1 \leq t' \leq
    t_{\max}} \,\land\, {\card{(\restr{S}{[t+1,t']})}\! \geq\! u} \}.\\
\end{array}
\]
}
\begin{definition}[Tuple-based window]
\label{def:tuple-based-window}
Let~${S=(T,\intpr)}$ be a stream and~${t \in T}$. Moreover, let~${\ell,
  u \in \bbN}$,~${T_\ell=[t_\ell,t]}$ and~${T_u=[t\!+\!1,t_u]}$,
where~$t_\ell$ and~$t_u$ are the tuple time bounds. The
\emph{tuple-based window with counts~$(\ell,u)$ of~$S$ at time~${t}$} is
defined by
\[
  w^{\# \ell, u}(S,t)=(T',\restr{\intpr'}{T'}),~\text{where}~T'= [t_\ell,t_u],~\text{and}
\]
\[v'(t') = \left\{
\begin{array}{ll}
  v(t') & \text{for all}~t' \in T' \setminus \{t_\ell,t_u\}\\
  v(t') & \text{if}~t'=t_\ell~~\text{and}~~\card{(\restr{S}{T_\ell})}\leq \ell\\
  X_\ell & \text{if}~t'=t_\ell~~\text{and}~~\card{(\restr{S}{T_\ell})}> \ell\\
  v(t') & \text{if}~t'=t_u~~\text{and}~~\card{(\restr{S}{T_u})}\leq u\\
  X_u   & \text{if}~t'=t_u~~\text{and}~~\card{(\restr{S}{T_u})}> u
\end{array}
\right.
\]
where~${X_q \subseteq \intpr(t_q)}$,~${q \in \{\ell,u\}}$, such
that~${\card{(T_q,\restr{\intpr'}{T_q})}=q}$.
\end{definition}
Note that the tuple-based window is unique only if for both~${q \in
  \{\ell,u\}}$,~${\intpr'(t_q)=\intpr(t_q)}$, i.e., if all atoms at the
endpoints of the selected interval are retained.
There are two natural possibilities to enforce the uniqueness of a
tuple-based window.
First, if there is a total order over all atoms, one can give a
deterministic definition of the sets~$X_q$ in
Def.~\ref{def:tuple-based-window}.
Second, one may omit the requirement that \emph{exactly}~$\ell$ tuples
of the past, resp.~$u$ tuples of the future are contained in the window,
but instead demand the substream obtained by the smallest
interval~$[t_\ell,t_u]$ containing \emph{at least}~$\ell$ past and~$u$
future tuples.
Note that this approach would simplify the definition to~${w^{\# \ell,
    u}(S,t)=(T',\restr{\intpr}{T'})}$, requiring only to
select~${T'=[t_\ell,t_u]}$.
We abbreviate the usual tuple-based window operator~$w^{\# \ell,0}$,
which looks only into the past, by~$w^{\# \ell}$. Similarly,~$w^{\# +u}$
stands for~$w^{\# 0,u}$.
\begin{example}[cont'd]
  To get the last~$3$ appearances of trams or buses from stream~$S$ in
  Example~\ref{ex:stream} at time point~$11$, we can apply a tuple-based
  window with counts~$(3,0)$. The application~$w^{\#3}(S,11)$ can lead
  to two possible windows~${(T',\intpr'_1)}$ and~${(T',\intpr'_2)}$,
  where~${T'=[2,11]}$, and
  \begin{align*}
    \intpr'_1 & =\{2 \mapsto \{\tram(a,p_1)\},
    8 \mapsto \{\tram(d,p_2)\},
    11 \mapsto \{\bus(e,p_2)\}\},\smallmathnl
    \intpr'_2 & =\{2 \mapsto \{\bus(c,p_1)\}, 8 \mapsto
    \{\tram(d,p_2)\}, 11 \mapsto \{\bus(e,p_2)\}\}.
  \end{align*}
  The two interpretations differ at time point~$2$, where
  either~$\tram(a,p_1)$ or~$\bus(c,p_1)$ is picked to complete the
  collection of~$3$ tuples.
\end{example}
The CQL syntax for the tuple-based window is~\texttt{[Rows n]}, which
corresponds to~$w^{\# n}$. Note that in CQL a single stream contains
tuples of a fixed schema. In the logic-oriented view, this would
translate to having only one predicate. Thus, applying a tuple-based
window on a stream in our sense would amount to counting tuples across
different streams. To enable counting of different predicates in
separation, we introduce a general form of partition-based windows.

The partition-based window CQL applies a tuple-based window function on
substreams which are determined by a sequence of attributes. The
syntax~\mbox{\texttt{[Partition By A1,...,Ak Rows N]}} means that tuples
are grouped into substreams by identical values~${a_1,\dots,a_k}$ of
attributes~\texttt{A1},\dots,~\texttt{Ak}. From each substream,
the~\texttt{N} tuples with the largest timestamps are returned.

Here, we have no notion of attributes. Instead, we employ a general
total \emph{index function}~${\idx : \GroundAtoms \rightarrow I}$ from
ground atoms to a finite \emph{index set}~${I \subseteq \bbN}$, where
for each~${i \in I}$ we obtain from a stream~${S=(T,\intpr)}$ a
substream~${\idx_i(S) =(T,\intpr_i)}$ by taking~${\intpr_i(t)=\{ a \in
  \intpr(t) \mid \idx(a) = i\}}$. Moreover, we allow for individual tuple
counts~${n(i)=(\ell_i,u_i)}$ for each substream~$S_i$.
\begin{definition}[Partition-based window]
  \label{def:partitioned-based-window}
  Let~${S=(T,\intpr)}$ be a stream, ${\idx:
    \GroundAtoms \rightarrow I \subseteq \bbN}$,
  an index function, and for all~${i \in
    I}$ let~${n(i)=(\ell_i, u_i) \in \bbN \times \bbN}$
  and~$S_i=\idx_i(S)$. Moreover, let~${t \in T}$ and~${w^{\#
      \ell_i,u_i}(S_i,t)=([t^\ell_i,t^u_i],\intpr'_i)}$ be the
  tuple-based window of counts~${(\ell_i,u_i)}$ of~${S_i}$ at time~$t$.
  Then, the \emph{partition-based window of counts~$\{(\ell_i,u_i)\}_{i
      \in I}$ of~$S$ at time~$t$ relative to~$\idx$} is defined by
  \[
  w^{\#n}_{\idx}(S,t)=(T',\intpr'),~\text{where}~T'=[\min_{i \in
    I}{t^\ell_i}, \max_{i \in I} t^u_i],
  \]
  and~${\intpr'(t')=\bigcup_{i \in I}\intpr'_i(t')}$ for all~${t' \in
    T'}$.
\end{definition}
Note that, in contrast to schema-based streaming approaches, we have
multiple kinds of tuples (predicates) in one stream. Whereas other
approaches may use tuple-based windows of different counts on separate
streams, we can have separate tuple-counts on the corresponding
substreams of a partition-based window on a single stream.
\begin{example}[cont'd]\label{ex:partition-based-window}
  Suppose we are interested in the arrival times of the last~$2$ trams,
  but we are not interested in buses. To this end, we construct a
  partition-based window~${w^{\# n}_{\idx}}$ as follows.
  We use index set~${I=\{1,2\}}$, and~${\idx(p(X,Y))=1}$
  iff~${p=\tram}$. For the counts in the tuple-based windows of the
  substreams, we use~${n(1)=(2,0)}$ and~${n(2)=(0,0)}$.
  We obtain the substreams
  \begin{align*}
    S_1 & =([2,13],\{2\mapsto\{\tram(a,p_1)\},8\mapsto\{\tram(d,p_2)\}\}),~\text{and}\smallmathnl
    S_2 & =([2,13],\{2\mapsto\{\bus(c,p_1)\},11\mapsto\{\bus(e,p_2)\}\}),
  \end{align*}
  and the respective tuple-based windows
  \begin{align*}
    w^{\#2}(S_1,13) &=([2,13],\{2\mapsto\!\{\tram(a,p_1)\},8\mapsto\!\{\tram(d,p_2)\}\}),~\text{and} \smallmathnl
    w^{\#0}(S_2,13) &=([13,13],\emptyset).
  \end{align*}
  Consequently, we get~${w_{\idx}^{\#n}(S,13)=([2,13],\intpr')}$, where~$\intpr'$ is
  \[ \{  2\mapsto\{\tram(a,p_1)\},8\mapsto\{\tram(d,p_2)\}. \]
\end{example}
\section{Reasoning over Streams \ifdraft(1-1.5)\fi}
\label{sec:stream-reasoning}
We are now going to utilize the above definitions of streams and windows
to formalize a semantics for stream reasoning.
\subsection{Stream Semantics}
Towards rich expressiveness, we provide different means to relate
logical truth to time.
Similarly as in modal logic, we will use operators~$\Box$ and~$\Diamond$
to test whether a tuple (atom) or a formula holds all the time,
respectively sometime in a window. Moreover, we use an~\emph{exact
  operator}~$@$ to refer to specific time points. To obtain a window of
the stream, we employ \emph{window operators}~$\window_i$.
\begin{definition}[Formulas  $\Formulas_k$]\label{def:formulas}
The set $\Formulas_k$ of  \emph{formulas (for $k$ modalities)} is
defined by the grammar
\smallskip

\centerline{$\alpha ::= a \mid \neg \alpha \mid \alpha \land \alpha \mid \alpha
  \lor \alpha \mid \alpha \rightarrow \alpha \mid \Diamond \alpha \mid
  \Box \alpha \mid @_t \alpha \mid \window_i \alpha$} %

\smallskip

\noindent where $a$ is any atom in $\cA$, $i \in\{1, \ldots k\}$, and $t \in
 \bbN\,{\cup}\, \TimeVariables$.
\end{definition}
We say a formula~${\alpha}$ is \emph{ground}, if all its atoms are ground
and for all occurrences of form~${@_t\beta}$ in~${\alpha}$ it holds
that~${t \in \bbN}$.
In the following semantics definition, we will consider the input stream
(\emph{urstream}) which remains unchanged, as well as dynamic substreams
thereof which are obtained by (possibly nested) applications of window
functions. To this end, we define a \emph{stream choice} to be a
function that returns a stream based on two input
streams.%
Two straightforward stream choices are~$ch_i$, for~${i \in \{1,2\}}$,
defined by~${ch_i(S_1,S_2)=S_i}$.
Given a stream choice~$ch$, we obtain for any window function~$w$ an
\emph{extended window function}~$\extwf$
by~${\extwf(S_1,S_2,t)=w(ch(S_1,S_2),t)}$ for all~${t \in \bbN}$. We
say~$\extwf$ is the \emph{extension of~$w$ (due to~$ch$)}.
\begin{definition}[Structure]\label{def:structure}
  Let~$S_M=(T,\intpr)$ be a stream,~${I
    \subseteq \bbN}$ a finite index set and let~$\windowf$ be a
  function that maps every~${i \in I}$ to an extended window
  function. The triple~${M=\langle T,\intpr,\windowf \rangle}$ is
  called a \emph{structure} and~${S_M}$ is called the \emph{urstream} of~$M$.
\end{definition}
We now define when a ground formula holds in a structure.
\begin{definition}[Entailment]
  Let~${M=\langle T,\intpr,\windowf \rangle}$ be a structure. For a
  substream~${S=(T_\ssS,\intpr_\ssS)}$ of~$(T,\intpr)$, we define the
  \emph{entailment} relation~$\entails$ between~${(M,S,t)}$ and
  formulas. Let~${t \in T}$,~${a \in \GroundAtoms}$, and~${\alpha, \beta
    \in \Formulas_k}$ be ground formulas and
  let~${\extwf_i=\windowf\!(i)}$. Then,
  \[
  \begin{array}{l@{\quad \text{iff}\quad}l}
    M,S,t \entails a  & a \in \intpr_\ssS(t)\,,\\%, \text{ for all } a \in \GroundAtoms\\
    M,S,t \entails \neg\alpha & M,S,t \notentails \alpha,\,\\
    M,S,t \entails \alpha \land \beta & M,S,t \entails \alpha~~\text{and}~~M,S,t \entails \beta,\,\\
    M,S,t \entails \alpha \lor \beta & M,S,t \entails \alpha~~\text{or}~~M,S,t \entails \beta ,\,\\
    M,S,t \entails \alpha \rightarrow \beta & M,S,t \notentails \alpha~~\text{or}~~M,S,t \entails \beta,\,\\
    M,S,t \entails \Diamond \alpha & M,S,t' \entails
    \alpha~~\text{for some}~~t'\! \in T_\ssS ,\,\\
    M,S,t \entails \Box \alpha & M,S,t' \entails \alpha~~\text{for all}~~t'\! \in T_\ssS\,,\\
    M,S,t \entails @_{t'} \alpha & M,S,t' \entails \alpha~~\text{and}~~t'\! \in T_\ssS\,,\\
    M,S,t \entails \window_i \alpha & M,S',t \entails \alpha~~\text{where}~~S'=\extwf_i(S_M,S,t).\\
  \end{array}
  \]
\end{definition}
If~$M,S,t \entails \alpha$ holds, we say~$(M,S,t)$
\emph{entails}~$\alpha$.
Intuitively,~$M$ contains the urstream~$S_M$ which remains unchanged
and~$S$ is the currently considered window. An application of a window
operator~$\window_i$ utilizes the extended window~$\windowf\!(i)$ which
can take into account both the urstream~$S_M$ and the current window~$S$
to obtain a new view, as we will discuss later.
The operators~$\Diamond$ and~$\Box$ are used to evaluate whether a
formula holds at some time point, respectively at all time points in the
timeline~$T_\ssS$ of~$S$. The operator~$@_t$ allows to evaluate whether
a formula holds at a specific time point~$t$ in~$T_\ssS$.
\begin{example}[cont'd]
  \label{ex:entailment}
  Let~${M=\langle T,\intpr,\windowf\rangle}$, where~${S_M=(T,\intpr)}$
  is the stream~$S$ from Example~\ref{ex:stream}
  and~${\windowf\!(1)=\extwf^5}$, i.e., the extension of~$w^5$ of
  Example~\ref{ex:time-based-windows} due to~$ch_2$. Consider the
  following formula:
  \[ \alpha=\window_1(\Diamond\tram(d,p_2)\land\Diamond\bus(e,p_2)) \]
  We verify that~${M,S_M,11 \entails \alpha}$ holds.
  First, the window operator~$\window_1$ selects the
  substream~${S'=(T_{\ssS'},\intpr')}$, where~${T_{\ssS'}=[6,11]}$
  and~${\intpr'=\restr{\intpr}{T'}=\{8\mapsto\{\tram(d,p_2)\},
    11\mapsto\{\bus(e,p_2)\}\}}$. Next, to see that~${(M,S',11)}$
  entails~${\Diamond\tram(d,p_2)\land\Diamond\bus(e,p_2)}$, we have to
  find time points in the timeline~${T_{\ssS'}}$ of the current
  window~$S'$, such that~${\tram(d,p_2)}$ and~${\bus(e,p_2)}$ hold,
  respectively. Indeed, for~$8$ and~$11$, we
  have~${M,S_1,8\entails\tram(d,p_2)}$
  and~${M,S_1,11\entails\bus(e,p_2)}$.~
\end{example}
\subsection{Queries}
\label{sec:queries}
We are now going to define the semantics of queries over streams.
\begin{definition}[Query] Let~${S=(T,\intpr)}$ be a stream,~${u \in T
    \cup \TimeVariables}$ and let~${\alpha}$ be a
  formula. Then~${\newquery{\alpha}{u}}$ denotes a~\emph{query
    (on~$S$)}. We say a query is \emph{ground}, if~$\alpha$ is ground
  and~$u \in T$, else \emph{non-ground}.
\end{definition}
For the evaluation of a ground query~$\alpha[t]$ we will use~$M,S_M,t \entails \alpha$.
To define the semantics of non-ground queries, we need the notions of
assignments and substitution. A \emph{variable assignment}~$\varAs$ is a
mapping~${\Variables\rightarrow\Constants}$ from variables to
constants. A \emph{time variable assignment}~$\timeAs$ is a
mapping~${\TimeVariables\rightarrow\bbN}$ from time variables to time
points. The pair~${(\varAs,\timeAs)}$ is called a \emph{query
  assignment}.
Table~\ref{tab:substitution} defines the
\emph{substitution}~$\querySubst$ based on query
assignment~${(\varAs,\timeAs)}$, where ${\alpha,\beta\in\Formulas_k}$.

\begin{toptable}
  \begin{center}
  \begin{tabular}{l@{~}l@{\quad}l}
    \multicolumn{2}{l}{\textsc{Definition}} & \textsc{Scope}\\[0.2ex]
    ${\querySubst(t)}$ & ${=~ t }$ & time points~${t \in \bbN}$\\
    ${\querySubst(u)}$ & ${=~ \tau(u)}$ & time variables~${u \in
      \TimeVariables}$\\
    ${\querySubst(c)}$ & ${=~ c}$ & constants~${c \in \Constants}$\\
    ${\querySubst(v)}$ & ${=~ \sigma(v)}$ & variables~${v \in
      \Variables}$\\
    \multicolumn{2}{l}{${\querySubst(p(t_1,\dots,t_n)) =}$} & predicates~${p
      \in \Predicates}$ and terms~${t_i \in \Terms}$
    \\
    \multicolumn{2}{l}{${~~~~~p(\querySubst(t_1),\dots,\querySubst(t_n))}$} & %
    \\
  ${\querySubst(\alpha\, \mb{b}\, \beta))}$ & ${=\querySubst(\alpha)\,\mb{b}\,\querySubst(\beta)}$ & ${\mb{b}\in\{\land,\lor,\rightarrow\}}$\\
  ${\querySubst(\mb{u} \alpha)}$ & ${=\mb{u}\querySubst(\alpha)}$ & ${ \mb{u}\in\{\neg,\Diamond,\Box\}\cup\{\window_i\}_{i \in \bbN}}$\\
  ${\querySubst(@_u \alpha)}$ & ${=@_{t} \querySubst(\alpha)}$ & ${@_u \alpha}$;~~${t=\querySubst(u)}$\\
  ${\querySubst(\newquery{\alpha}{u})}$ & ${=\newquery{\querySubst(\alpha)}{\querySubst(u)}}$ & queries~${\newquery{\alpha}{u}}$
\end{tabular}
\caption{Definition of substitution~$\querySubst$ based on query
  assignment~$(\varAs,\timeAs)$}
\label{tab:substitution}
\end{center}
\vspace{-0.9cm} %
\end{toptable}
Let~${q=\newquery{\alpha}{u}}$ be a query on~${S=(T,\intpr)}$.  We say a
substitution~$\querySubst$ \emph{grounds}~$q$, if~$\querySubst(q)$ is
ground, i.e., if~$\querySubst$ maps all variables and time variables
occurring in~$q$. If, in addition,~${\timeAs(x) \in T}$ for every time
variable~${x \in \TimeVariables}$ occurring in~$q$, we say~$\querySubst$
is~\emph{compatible} with~$q$.
\begin{definition}[Answer] The \emph{answer}~${?q}$ to a
  query~${q=\newquery{\alpha}{t}}$ on~$S$ is defined as follows.
  If~$q$ is ground, then~${?q=\yes}$ if~${M,S_M,t \entails q}$ holds, and~${?q=\no}$
  otherwise. If~$q$ is non-ground, then
\[ {?q = \{ (\varAs,\timeAs) \mid \querySubst~\text{is compatible
    with}~q~\text{and}~?\!\querySubst(q)=\yes \}}. \]
\end{definition}
That is, the answer to a non-ground query is the set of query
substitutions such that the obtained ground queries hold.
\begin{example}[cont'd]
  \label{ex:non-ground}
  We formalize the queries of Ex.~\ref{ex:traffic-scenario} as follows:
  \begin{align*}
    q_1 & = \window_1(\Diamond\tram(X,P)\land\Diamond\bus(Y,P))[u]\smallmathnl
    q_2 & = \window_1\Diamond(\tram(X,P)\land\bus(Y,P))[u]
  \end{align*}
  The query~${q=\window_1\Diamond(\tram(a,p_1)\land\bus(c,p_1))[t]}$ is
  ground iff~${t \in \bbN}$ and $?q=\yes$ iff~${t \in [2,7]}$.
  We evaluate~$q_1$ on
  structure~$M$ of Ex.~\ref{ex:entailment}:
  \begin{align*}
    M,S_M,t &\entails\window_1(\Diamond\tram(a,p_1)\land\Diamond\bus(c,p_1))~~\text{for
      all}~~{t\in [2,7]}\smallmathnl
    M,S_M,t & \entails\window_1(\Diamond\tram(d,p_2)\land\Diamond\bus(e,p_2))~~\text{for all}~~{t\in[11,13]}
  \end{align*}
  Thus, the following set of substitutions is the answer to~$q_1$
  in~$M$:
\vspace{-1.0ex}
\begin{displaymath}
  \begin{array}{r@{~}l}
    ?q_1 = & \{(\{X\!\mapsto\! a,Y\!\mapsto\! c,P\!\mapsto\! p_1\},\{u\!\mapsto\! t\}) \mid t\in[2,7]\}\, \cup \\
    & \{(\{X\!\mapsto\! d,Y\!\mapsto\! e,P\!\mapsto\! p_2\},\{u\!\mapsto\! t\}) \mid
    t\in[11,13]\}
  \end{array}
\end{displaymath}
\end{example}
\leanparagraph{Exact time reference}
With the operator~$@_t$ we can ask whether a formula holds at a specific
time point~$t$. In its non-ground version, we can utilize this operator
for the selection of time points.
\begin{example}[cont'd]
  \label{ex:exact-time-reference}
  Let~${\alpha = tram(X,P) \land \bus(Y,P)}$. For each of the
  queries ${@_U \alpha [13]}$ and~${\alpha[U]}$, the time assignments
  for~$U$ in the answers will map to time points when a tram and a bus
  arrived simultaneously at the same stop. In both cases, the single
  answer is~${ (\{X \mapsto a, Y \mapsto c, P \mapsto p_1\},\{U \mapsto
    2\})}$.
  Note that omitting~$@_U$ in the first query would give an empty
  answer, since the subformula~$\alpha$ does not hold at time
  point~$13$.
\end{example}
We observe that the operator~${@}$ allows to replay a historic query. At
any time~${t'>t}$, we can ask~${@_t\alpha[t']}$ to simulate a previous
query~${\alpha[t]}$.

\leanparagraph{Nested windows}
Typically, window functions are used exclusively to restrict the
processing of streams to a recent subset of the input. In our %
view, window functions provide a flexible means to reason over
temporally local contexts within larger windows.
For these nested windows we carry both~$M$ and~$S$ for the entailment
relation.
\begin{example}[cont'd]
  \label{ex:nested-windows}
  Consider the following additional query $(q_3)$: At which
  stops~$P$, for the last~$2$ two trams~$X$, did a bus~$Y$ arrive
  within~$3$ minutes?
  To answer $(q_3)$ at time point~$13$, we ask
  \begin{displaymath}
    q_3 = \window_1\Box(\tram(X,P)\rightarrow\window_2\Diamond\bus(Y,P))[13].
  \end{displaymath}
  For~$\window_1$, we can use the extension~$\extwf^{\# n}_{\idx}$ of
  the partition-based window~$w^{\# n}_{\idx}$ of
  Example~\ref{ex:partition-based-window}.
  Applying~${\windowf\!(1)}$ on the stream~${S=(T,\intpr)}$ in the previous
  examples yields~${S'=(T',\intpr')}$, where~${T'=[2,13]}$
  and~${\intpr'=\{2 \mapsto \{\tram(a,p_1)\}, 8 \mapsto
    \{\tram(d,p_2)\}\}}$.
  That is, after applying this window, the current window~$S'$
  no longer contains information on buses.
  Consequently, to check whether a bus came in both cases within~$3$
  minutes, we must use the urstream~$S_M$. Thus, the second extended
  window~${\windowf\!(2)=\extwf^{+3}}$ is the extension of the
  time-based window~$w^{+3}$, which looks~$3$ minutes into the future,
  due to the stream choice~$ch_1$. Hence,~$\extwf^{+3}$ will create
  a window based on~$S_M$ and not on~$S'$.
  The two time points in~$T'$ where a tram appears are~$2$ and~$8$,
  with~$P$ matching~$p_1$ and~$p_2$,
  respectively. Applying~$\windowf\!(2)$ there yields
  the streams~${S''_2=(T''_2,\intpr''_2)}$
  and~${S''_8=(T''_8,\intpr''_8)}$, where
  \begin{align*}
    T''_2 &= [2,5], & \intpr''_2 &= \{ 2 \mapsto \{ \tram(a,p_1), \bus(c,p_1) \} \},~\text{and}\\
    T''_8 &= [8,11], & \intpr''_8 &= \{ 8 \mapsto \{ \tram(d,p_2) \}, 11
    \mapsto \{ \bus(e,p_2) \} \}.
  \end{align*}
  In both streams, we find a time point with an
  atom~$\bus(Y,p_j)$ with the same stop~$p_j$ as the tram. Thus,
  in both cases the subformula~${\Diamond\bus(Y,P)}$ is satisfied
  and so the
  implication~${\tram(X,P)\rightarrow\window_2\Diamond\bus(Y,P)}$ holds
  at every point in time of the stream selected by~$\window_1$. Hence,
  the answer to the query is
  \[
  \begin{array}{r@{}l}
    ?q_3 = \{ & \{ (X\mapsto a, Y\mapsto c, P\mapsto p_1\},\emptyset)\}, \\
    & \{ (X\mapsto d, Y\mapsto e, P\mapsto p_2\},\emptyset)\}\}.
  \end{array}
  \]
\end{example}
\section{Discussion and Related Work \ifdraft(1.5-1.75p)\fi}
\label{sec:discussion}
In this section we discuss the relationship of this ongoing work with
existing approaches from different communities.

\leanparagraph{Modal logic} The presented formalism employs operators~$\Diamond$ and~$\Box$ as in modal
logic~\cite{Blackburn01}. Also, the definition of entailment uses
a structure similar to Kripke models for multi-modal logics. However, instead
of static accessibility relations, we use window functions which take
into account not only the worlds (i.e., the time points) but
also the interpretation function. To our best knowledge, window operators
have been considered neither in modal logics nor temporal logics.

\leanparagraph{CQL}
By extending SQL to deal with input streams, CQL queries are
evaluated based on three sets of operators:
\begin{enumerate}[(i)]
\item\label{cql-op-s2r} \emph{Stream-to-relation} operators apply window
  functions to the input stream to create a mapping from execution times
  to bags of valid tuples (w.r.t. the window) without timestamps. This
  mapping is called a relation.
\item\label{cql-op-r2r} \emph{Relation-to-relation} operators allow for
  modification of relations similarly as in relational algebra,
  respectively SQL.
\item\label{cql-op-r2s} \emph{Relation-to-stream} operators convert
  relations to streams by directly associating the timestamp of the
  execution with each tuple (RStream). The other operators
  IStream/DStream, which report inserted/deleted tuples, are derived from
  RStream.
\end{enumerate}
The proposed semantics has means to capture these operators:
\begin{enumerate}[(i)]
\item\label{capture-cql-1} The window operators~$\window_i$ keep the
  timestamps of the selected atoms, whereas the stream-to-relation
  operator discards them. The CQL query for tuple~$x$ thus corresponds
  to a query~$\Diamond x$ of the present setting.
  A stream in CQL belongs to a fixed schema. As noted earlier, this
  corresponds to the special case with only one predicate.
  CQL's partition-based window is a generalization of the tuple-based
  window defined there. In turn, the presented partition-based window
  generalizes the one of CQL. %
\item\label{capture-cql-2} Some relational operators can essentially be
  captured by logical connectives, e.g., the join by conjunction. Some
  operators like projection will require an extension of the formalism
  towards rules. Moreover, we did not consider arithmetic operators
  and aggregation functions, which CQL inherits from SQL.
\item\label{capture-cql-3} The answer to a non-ground query~$\alpha[u]$
  is a set of query assignments~${(\varAs,\timeAs)}$. To capture the
  RStream of CQL, we can group these assignments by the time
  variable~$u$.
\end{enumerate}

\begin{example}
  \label{ex:capture-cql}
  Queries ($q_1$) and ($q_2$)
  from Example~\ref{ex:traffic-scenario} can be expressed in CQL.
  We assume that both streams have the attributes~$X$ and~$P$,
  corresponding to the first, respectively second argument of
  predicates~$\tram$ and~$\bus$.
  For~($q_1$), we can use:
\begin{verbatim}
  SELECT * FROM tr [RANGE 5], bus [RANGE 5]
  WHERE tr.P = bus.P
\end{verbatim}
  On the other hand,~($q_2$) needs two CQL queries.
\begin{verbatim}
  SELECT * AS tr_bus FROM tr [NOW], bus [NOW]
  WHERE tr.P = bus.P
\end{verbatim}
  \vspace{-12pt}
\begin{verbatim}
  SELECT * FROM tr_bus [RANGE 5]
\end{verbatim}
  Here, the first query produces a new stream that contains only
  simultaneous tuples and the second one covers the range of~$5$
  minutes.
\end{example}
Traditionally, stream reasoning approaches use \emph{continuous
  queries}, i.e., repeated runs of queries with snapshot semantics to
deal with changing information. In this work, we go a step further and
enable reasoning over streams within the formalism itself by means of
nested windows.
One can only mimic this feature with CQL's snapshot semantics when
timestamps are part of the schema and explicitly encoded.
Likewise, queries to future time points can be emulated in this way, as
the next example shows.
\begin{example}[cont'd]
  In Example~\ref{ex:nested-windows}, we considered bus arrivals
  within~$3$ minutes after the last~$2$ trams.
  In CQL, such a query is not possible on the assumed schema. However,
  by adding a third attribute~\texttt{TS} that carries
  the timestamps to the schema, the following CQL query yields the same results.
\begin{verbatim}
  SELECT * FROM tr  [ROWS 2],
                bus [RANGE UNBOUNDED]
  WHERE tr.P = bus.P AND bus.TS - tr.TS <= 3
\end{verbatim}
  Note that we need no partition-based window here, since trams and
  buses arrive from different input streams. Moreover, we must use the
  unbounded window for buses to cover nesting of windows in~($q_3$)
  because windows in CQL are applied at query time and not the time
  where a tram appearance is notified.
\end{example}
Furthermore, nested CQL queries and aggregation inherited from SQL are
promising to mimic the behavior of operator~$\Box$. With according
rewriting, CQL eingines like STREAM~\cite{ArasuBBDIRW03} could be used
to realize the proposed semantics.

\leanparagraph{SECRET}
In~\cite{DindarTMHB13} a model called SECRET is proposed to analyze the
execution behavior of different stream processing engines (SPEs) from a
practical point of view.
The authors found that even the outcome of identical, simple queries
vary significantly due to the different underlying processing models.
There, the focus is on
understanding, comparing and predicting the \emph{behaviour of
  engines}. In contrast, we want to provide means that allow for a
similar analytical study for the \emph{semantics} of stream reasoning
formalisms and engines.
The two approaches are thus orthogonal and
can be used together to compare stream reasoning engines based on
different input feeding modes as well as different reasoning
expressiveness.

\leanparagraph{Reactive ASP}
The most recent work related to expressive stream reasoning with
rules~\cite{GebserGKOSS2012} is based on Reactive
ASP~\cite{GebserGKS11}. This setting introduces logic programs that
extend over time. Such programs have the following components. %
Two components~$P$ and~$Q$
are parametrized with a natural number~$t$ for time points. In addition,
a basic component~$B$ encodes background knowledge that is not time-dependent.
Moreover, sequences of pairs of arbitrary logic programs~$(E_i,F_j)$, called \emph{online
progression} are used. While~$P$ and~$E_i$ capture accumulated knowledge,~$Q$
and~$F_j$ are only valid at specific time points. Compared to
reactive ASP,
  our semantics has no mechanism for accumulating programs, and
  we take only streams of atoms/facts, but no background theories.
Therefore, a framework based on idealized semantics with extension to
rules should be able to capture a fragment of reactive ASP where~$P$
and~$F_j$ are empty and~$E_i$ contains only facts. The foreseeable
conversion can be as follows: convert rules in~$Q$ by applying an
unbounded window on all body atoms of a rule, using~$@_t$ to query the
truth value of the atoms at time point~$t$. Then, conclude the head to be
true at~$t$ and feed facts from~$E_i$ to the input stream~$S$.

\leanparagraph{StreamLog}
Another logic-based approach towards stream reasoning is \mbox{StreamLog}~\cite{Zaniolo12}. It makes use of Datalog and introduces
\emph{temporal predicates} whose first arguments are timestamps. By
introducing \emph{sequential programs} which have syntactical
restrictions on temporal rules, \mbox{StreamLog} defines
\emph{non-blocking negation} (for which Closed World Assumption can be
safely applied) that can be used in recursive rules in a stream
setting. Since sequential programs are locally stratified, they have
efficiently computable perfect (i.e., unique) models. Similar to
capturing a fragment of Reactive ASP, we can capture \mbox{StreamLog} by
converting temporal atoms~$p(t,x_1,\dots,x_n)$ to expressions~$@_t
p(x_1,\dots,x_n)$ and employing safety conditions to rules to
  simulate non-blocking negation. Moreover, we plan for having weaker
notions of negation that might block rules but just for a bounded number
of time points to the future.

\leanparagraph{ETALIS} The ETALIS system \cite{arfs2012} aims at adding
expressiveness to Complex Event Processing (CEP). It provides a
rule-based language for pattern matching over event streams with
\emph{declarative monotonic semantics}. Simultaneous events are not
allowed and windows are not regarded as first-class objects in the
semantics, but they are available at the system implementation
level. Tuple-based windows are also not directly supported.
Furthermore,
nesting of windows is not possible within the language, but
it can be emulated with multiple rules as in CQL. On the other hand,
ETALIS models complex events with time intervals and has operators
to express temporal relationships between events.
\section{Conclusion \ifdraft(0.25p)\fi}
\label{sec:conclusion}
We presented a first step towards a theoretical foundation
for (idealistic)  semantics of stream reasoning formalisms. %
Analytical tools to characterize, study and compare logical aspects of
stream engines %
have been missing. To fill this gap, we provide a framework to reason over
streaming data with a fine-grained
control over relating the truth of tuples with their occurrences in time. It
thus, e.g., allows to capture various kinds of window applications on data streams.
We discussed the relationship of the proposed formalism with exsisting
approaches, namely CQL, SECRET, Reactive ASP, StreamLog, and ETALIS.

Next steps include extensions of the framework to formally capture
fragments of existing approaches. Towards more
advanced reasoning features like recursion and non-monotonicity, we aim
at a rule-based semantics on top of the presented core.
Furthermore, considering intervals of time as references is
an interesting research issue.
To improve practicality (as a tool for formal and experimental analysis)
one might also develop an operational characterization of the
framework. In a longer perspective, along the same lines
with~\cite{Brewka13}, we aim at a formalism for stream reasoning in
distributed settings across heterogeneous nodes having potentially
different logical capabilities.%

\end{document}